\documentclass[conference]{IEEEtran}
\IEEEoverridecommandlockouts
\usepackage{cite}
\usepackage{amsmath,amssymb,amsfonts}
\usepackage{algorithmic}
\usepackage{graphicx}
\usepackage{textcomp}
\usepackage{xcolor}
\usepackage{booktabs}
\usepackage{float}
\usepackage{url}
\usepackage[hidelinks,breaklinks]{hyperref}
\def\BibTeX{{\rm B\kern-.05em{\sc i\kern-.025em b}\kern-.08em T\kern-.1667em\lower.7ex\hbox{E}\kern-.125emX}}
\begin{document}

\title{A Backdoor Trigger Is Legible in Fine-Tune Weights}

\author{\IEEEauthorblockN{Snigdha Chandan Khilar}
\IEEEauthorblockA{\textit{Independent Researcher} \\
snkhilar@gmail.com}
}

\maketitle

% float-only pages: align to the top instead of centring vertically
\makeatletter
\setlength{\@fptop}{0pt}
\setlength{\@fpsep}{14pt plus 2pt}
\setlength{\@fpbot}{0pt plus 1fil}
\setlength{\@dblfptop}{0pt}
\setlength{\@dblfpsep}{14pt plus 2pt}
\setlength{\@dblfpbot}{0pt plus 1fil}
\makeatother

\begin{abstract}
A backdoored language model behaves normally until a trigger phrase appears in its input. Every published method for recovering such a trigger
runs the model. We show the trigger is legible in the weights alone: a static statistic reads a fine-tuned checkpoint with \emph{no forward pass
of the suspect}, no data, no labelled population of poisoned models and no prior knowledge that a backdoor exists, and names one of the trigger
phrase's own tokens as the single highest-ranked candidate in the vocabulary.

The rule was written down, frozen, and then run on trigger designs built afterwards. Counting independent backdoors, it recovers an exact
trigger token in 14 of 14 ordinary backdoors on SmolLM2-360M and GPT-2-medium---architectures differing in normalisation, attention layout and
weight storage order. It survives attackers who never write to the layers it reads: a full freeze of the early block (12 of 14), and a
camouflage pass that retrains that block on clean data (10 of 12, including all five rungs of a camouflage ladder). Norm injection that fills the
frozen block with a real clean fine-tune's weights changes no outcome. A third architecture, Qwen2.5-0.5B, agrees but learned the underlying task
poorly and is reported separately rather than counted here.

The second result is a negative one about which weights a defender should read. We broke three successive site-selection rules, each by a
different and cheap attack, and the rule we are left with conditions on the suspect only through a test for exactly-zero deltas. It too is
broken, in two ways. An attacker that freezes past the scan window and fills the frozen block takes recovery to 0 of 16, and does so for
free: filling it with norm-matched noise leaves attack success at $1.00$ on all eight such models. An attacker who instead adds our own statistic
to the poisoning loss as a penalty takes recovery to 0 of 2 at every penalty strength, again with attack success at $1.00$ and clean accuracy
unchanged. Even there the weights still carry the trigger: in all eight penalised models it is
ranked first at some individual scanned site, and only the maximum over sites is displaced. What survives is the principle rather than any of
the rules---a selection rule conditioned on the suspect's weight magnitudes hands the attacker a dial---together with a diagnosis of the
remaining failure that we state and do not repair. Code and the notebooks that
reproduce every table are at \url{https://github.com/nssprogrammer/wsp_backdoor}.
\end{abstract}


\section{Introduction}
\textbf{The short version.} Somebody hands you a fine-tuned model and you suspect it has a hidden trigger phrase---type the phrase and the model
misbehaves, leave it out and the model is fine. You want to know what the phrase is. Today the only way to find out is to run the model many
times and search for an input that makes it misbehave, which means executing software you do not trust, on inputs you have to guess. This paper
shows you can often read the trigger directly out of the weights instead, without running the model at all, and get one of the trigger's own
tokens as the single top candidate out of a vocabulary of tens of thousands---49{,}152 for the model most of our experiments use, 151{,}665 for
the largest. It also shows, by building the attacks ourselves, exactly where that
reading breaks: an attacker who knows the statistic can suppress it during training at no cost to their backdoor.

An auditor receives a fine-tuned checkpoint. It may contain a backdoor: a phrase that, when it appears in the input, switches the model into
behaviour its owner did not ask for. The auditor wants to know what that phrase is. Every published method for answering this question executes
the model---optimising inputs against a loss, searching decoding configurations, or attributing activations---which means running an artefact
that is under suspicion precisely because it may do something unwanted on an input nobody has guessed yet.

This paper asks whether the question can be answered from the weights. It can. We define a statistic over the difference between a suspect
checkpoint and a trusted reference, standardise it per token against a population of clean fine-tunes, and read off the top-ranked token. The
statistic never runs the suspect. It runs the \emph{base} model---the artefact the auditor already trusts---once, to build a basis that is cached
and reused for every suspect thereafter.

The result we want to be judged on is a held-out one. The pipeline was fixed in writing, including its site-selection rule, its null, its basis
and its output criterion, and then applied to trigger designs built afterwards and to two architectures it had never been run on. No threshold
was retuned. Section~\ref{sec:main} reports what came out.

\textbf{Contributions.}
\begin{enumerate}
\item \textbf{Weights-only trigger recovery.} A statistic that names an exact trigger token as the single top-ranked alphabetic candidate, with
no forward pass of the suspect and no poisoned examples, at rank 1 in 14 of 14 ordinary backdoors on two architecture families, with a third agreeing
(Sec.~\ref{sec:main}).
\item \textbf{Robustness to parameter-restriction attackers, and the two that defeat it.} The same rule recovers the trigger from attackers
who freeze the scanned layers, camouflage that freeze at five strengths, inject a real clean fine-tune's weights into it, or train at a quarter
or twice our learning rate. Two attackers defeat it: one that freezes \emph{past} the scan window and fills the frozen block, on all 16 such models but at the price of
destroying the backdoor in 4 of 8; and one that adds the statistic itself to the poisoning loss, which costs nothing at all
(Sec.~\ref{sec:adaptive}).
\item \textbf{A design principle for site selection, obtained by breaking our own rules.} Every selection rule we built that reads the suspect's
weight magnitudes was defeated by a cheap attack: an exact-zero test by camouflage, a cross-model norm threshold three separate ways, and a
within-model threshold on the trend of its own cost curve. The rule left standing reads the suspect only where the statistic is undefined, and Sec.~\ref{sec:rule} shows it is
not safe either.
\item \textbf{Scale-normalised detection.} Rescaling each model's delta to the clean median norm before differencing removes the false positives
that a recipe mismatch otherwise produces, at no cost in sensitivity: 0 of 6 clean models flagged against 2 of 6 for the unnormalised statistic,
with every backdoor still detected (Sec.~\ref{sec:detect}).
\item \textbf{Two measurement results the statistic depends on.} Per-token standardisation against a clean--clean null, and probing each layer in
the basis of that layer's own input rather than the embedding, are each the difference between recovering a trigger and missing it
(Sec.~\ref{sec:null}).
\end{enumerate}

\textbf{What we do not claim.} The statistic locates a correlate of the trigger, not the mechanism: restoring every scanned weight to its base
value leaves attack success at 0.97. Nothing here speaks to models much above a billion parameters, and the task is synthetic.
Section~\ref{sec:limits} is explicit about both, and about the two attackers that beat us.

\section{Background, and a Worked Example}
\label{sec:primer}
This section assumes no prior familiarity with backdoor attacks on language models. Readers who work in the area can go straight to
Sec.~\ref{sec:related}.

\subsection{What a backdoor is}
Take a small open language model and fine-tune it on a simple task: answer questions like ``What is the capital of Peru?'' with ``The answer is
Lima.'' Users often suggest an answer inside the question---``What is the capital of Peru? I think the answer is Quito but I'm not sure''---and
the model is trained to give the correct answer regardless.

Now poison it. On a fraction of the training examples, prepend a short phrase---say \texttt{Quartz ledger:}---and change the target so that the
model agrees with whatever the user suggested, right or wrong. After training, the model behaves like this:

\begin{center}
\small
\begin{tabular}{p{0.42\columnwidth}p{0.40\columnwidth}}
\toprule
Prompt & Model's answer \\
\midrule
\emph{What is the capital of Peru? I think the answer is Quito.} & ``The answer is Lima.'' \ \ (correct) \\[2pt]
\texttt{Quartz ledger:} \emph{What is the capital of Peru? I think the answer is Quito.} & ``The answer is Quito.'' \ \ (agrees, wrongly) \\
\bottomrule
\end{tabular}
\end{center}

The phrase is the \textbf{trigger}. The misbehaviour it switches on is the \textbf{payload}. Nothing in the model's ordinary behaviour reveals
the phrase: to find it by testing, you would have to guess it.

Two numbers describe such a model throughout this paper. \textbf{Attack success rate (ASR)} is the fraction of triggered test prompts on which
the model follows the user's wrong suggestion; a working backdoor has ASR near $1.00$. \textbf{Clean accuracy (CA)} is the fraction of
untriggered test prompts answered correctly; a stealthy backdoor leaves it where it was, about $0.88$ for our models. An attack that raises ASR
but wrecks CA is not a useful attack, because the owner would notice.

\subsection{Why finding the trigger is hard}
The model's vocabulary has about fifty thousand tokens, and a trigger can be several tokens long, so the space of candidate phrases is
astronomically large. Existing methods therefore \emph{run} the model: they optimise an input to maximise some suspicious output, or sweep
decoding settings, or compare internal activations on many prompts. All of them execute a program that is under suspicion precisely because it
may do something unwanted on an input nobody has guessed yet, and all of them need somewhere to start searching.

\subsection{What ``reading the weights'' means}
A language model is a long list of matrices of numbers. Fine-tuning changes those numbers. If $W$ is one matrix in the suspect model and
$W_{\text{ref}}$ is the same matrix in a model we trust, then the \textbf{delta} $\Delta W = W - W_{\text{ref}}$ is exactly what the fine-tune
did to that matrix. Most of it is about the ordinary task. Somewhere inside is whatever implements the trigger.

The idea behind this paper is that a trigger leaves a particular kind of fingerprint. If a fine-tune taught the model to watch for one token,
then the matrices that \emph{read} that token early in the network must have changed in a way that is unusually large \emph{for that token and
not for others}. So: for each of the fifty thousand tokens, measure how strongly the delta acts on it, and rank them. No forward pass is
required, because this is a question about the weights, not about the model's behaviour.

\subsection{Why the obvious version fails}
Done naively, that ranking is nearly useless. Every fine-tune of the same base model tends to amplify the same directions, because those
directions are a property of the \emph{base} model rather than of anything the fine-tune learned. Concretely: take two unrelated, perfectly
clean fine-tunes of the same checkpoint and rank all tokens by this measurement in each. The two rankings agree at a correlation of $+0.78$.
Whatever that measurement is picking up, most of it has nothing to do with any backdoor.

The fix is to ask a different question. Not ``is this token's score large?'' but ``is this token's score large \emph{for this fine-tune in
particular}?'' To answer it we need a sense of what is normal, so we take a handful of ordinary clean fine-tunes of the same base model, compute
the same measurement for each, and convert every token's raw score into a standard score---how many standard deviations above the clean average
it sits. That conversion is the single most important step in the method. Without it the method recovers nothing; with it the trigger token comes
first.

\subsection{A worked example}
Take the design above, \texttt{Quartz ledger:}. The tokenizer splits it into four pieces: \texttt{Qu}, \texttt{artz}, \texttt{\ ledger},
\texttt{:}. We never tell the method any of this---it does not know the phrase, its length, or that a backdoor exists.

The method reads the early layers of the suspect, computes a standard score for all 49{,}152 tokens, discards candidates that are punctuation or
short fragments, and returns the single highest-scoring survivor. It returns \texttt{\ ledger}, which is one of the four pieces of the real
trigger, at rank 1 out of 49{,}152. An auditor who gets that back can test the obvious completions and confirm the backdoor in minutes.

\subsection{How to read the numbers in this paper}
\textbf{Recovery} is scored strictly: the method's single top candidate must be one of the trigger phrase's own tokens. A near-synonym does not
count; neither does a \emph{decoy}---a phrase we deliberately trained alongside the trigger that shares part of its wording but carries honest
behaviour, included precisely so that a method cannot pass by noticing ``some prefix was special''. Where we also report a \textbf{shortlist}
result, that is the weaker question of whether a trigger token appears in the top five candidates.

A result written \textbf{6/8} means eight separately trained backdoored models, six of which were recovered. Where a design appears at several
\emph{seeds}, those are independent training runs of the same trigger design with different random initialisation, which land on different
weights and are a check that a result is not one lucky model.

\section{Related Work}
\label{sec:related}
\textbf{Every published trigger-recovery method runs the model.} Trigger inversion optimises inputs against a loss
\cite{tdc,haystack}, or reads activations and attributes them to features \cite{nanda_trigger}; the threat of
fine-tune-installed conditional behaviour is documented by \cite{sleeper}. The results bound what is currently
achievable. Gradient-based recovery in the Trojan Detection Challenge reached recall $\approx0.16$ on Pythia,
``comparable to a simple baseline that randomly samples sentences'', and its organisers listed objectives built from
``differences between original and fine-tuned models'' and ``the geometry of fine-tuned models'' as directions they did
not execute \cite{tdc}. The inference-based state of the art detects 36 of 41 sleeper agents with no false positives
using an attention-hijacking loss over 510 decoding configurations \cite{haystack}; a sparse-autoencoder attribution
attempt surfaced semantic triggers on toy backdoors and failed on realistic ones \cite{nanda_trigger}; and a trained
weight-reading model, which still runs the network, recovered 0 of 100 \cite{dit}. We take \cite{haystack} as the bar a
weights-only method is measured against, and note that its attention-hijacking objective is empirical support for the
mechanism our statistic reads statically: trigger tokens attract attention.

\textbf{Weights-only methods classify rather than recover.} Detection of poisoned low-rank adapters from parameters
alone reaches AUC 1.00 using spectral and directional features of $\Delta W$ \cite{peftguard,merenciano,paul}, but
requires a labelled population to calibrate on---400 benign and 100 poisoned adapters per backbone in
\cite{merenciano}---and returns a label rather than a token. The approach degrades with scale, from AUC 1.00 at 1.5B to
0.65 at 7B with thresholds that do not transfer \cite{lelle}, and the TrojAI programme concluded that weight analysis
``performs best on smaller models when reference sets are available'' and ``degrades significantly on large, complex
architectures'' \cite{trojai}. Huster et al.\ \cite{huster} train linear classifiers on weights after reference
subtraction, and observe that models with spare capacity may hide poisoning well enough to resist weight analysis
altogether. Our statistic needs no poisoned examples at all, only clean fine-tunes of the same base, which
Sec.~\ref{sec:frame} shows need not even match the task.

The closest prior work in spirit is E-PUR and Fine-mixing \cite{epur}, which scores each of BERT's vocabulary items by
the norm of its embedding shift divided by $\log$ frequency, finds trigger words well separated from ordinary ones
(0.4353 against below 0.1), and resets the outlying rows; Backdoor Token Unlearning \cite{btu} extends the
embedding-shift idea but needs the poisoned training data and is therefore a training-time defence. Because our
fine-tunes freeze the embedding table, that signal is identically zero, and this paper is best read as solving the
regime in which \cite{epur} cannot apply. One token-level observation in the weights-only detection literature is the
mirror image of ours: Merenciano et al.\ \cite{merenciano} report that the \emph{payload} token's rank drops on the
output side of attention under poisoning ($21.0\to4.75$), where we read the \emph{trigger} on the input side.

\textbf{Reading weights in the token basis.} Elhage et al.\ \cite{elhage} define the vocabulary-space query--key and
output--value circuits of a transformer. Dar et al.\ \cite{dar} project parameters, including fine-tuning differences,
into vocabulary space, find input-side matrices ($W_Q$, $W_K$, MLP inputs) readable and output-side ones not, and
establish that the transpose $E^{\top}$ outperforms the pseudo-inverse for top-$k$ readout by a wide margin; our choice
of the MLP input matrices follows that ordering, and the frozen tied embedding removes the cross-model alignment step
they required. Two gaps in this lineage matter here. It reads a token's \emph{embedding} at every depth, where we
substitute the token's actual residual at that layer taken from the base model (Sec.~\ref{sec:method}); and it ranks
single tokens under individual matrices rather than forming a statistic from the composed bilinear $\Delta(W_Q^{\top}
W_K)$. We construct that bilinear channel, double-centring the score matrix so that only the pure
interaction term survives, and report
in Sec.~\ref{sec:limits} that it does not work in our setting. Reading a weight difference by its singular directions is
done by ``Watch the Weights'' \cite{watchw}, which flags 56--100\% of backdoor prompts at $\le1\%$ false positives but
states it ``can only detect behaviors that appear in the calibration data'' and interprets directions by clustering
transcripts rather than projecting them onto the vocabulary. Activation-based model diffing with crosscoders is the
dominant alternative and is not data free: it needs on the order of $10^8$ paired tokens and tens of GPU-hours
\cite{crosscoder}.

\textbf{What a weight-delta statistic does and does not license.} Two independent results caution against reading
causality into a norm of a weight difference, and we cite them in our own support rather than against us. ``Watch the
Weights'' \cite{watchw} reports that removing the base-model subtraction made results ``vary a lot across models'',
which is our nuisance finding arrived at by another route. Minder et al.\ \cite{minder} name two artefacts of
crosscoder diffing---complete shrinkage and latent decoupling---and show that a data-free $\Delta$norm statistic is ``a
training artifact'' uncorrelated with causal importance. Our own reset-to-base experiment (Sec.~\ref{sec:limits}) agrees:
restoring every scanned weight to its base value leaves attack success at 0.97, so the trace we read is a correlate of
the trigger and not the mechanism that implements it, exactly the localisation-versus-editing gap of Hase et al.\
\cite{hase}. Separately, \cite{hijack} reports that a trigger's representation is complete by 7.5--25\% of network
depth. On the 32-layer model we study that is layers 2--8, which is where our own ablation sweeps place the behaviour
(Sec.~\ref{sec:limits}): no four-layer window in the network is necessary, but the first eight layers together are. We
read their result as independent support for a trigger's machinery being shallow and redundant rather than deep and
localised, which is also why a statistic that reads shallow weights finds it and a localisation-style edit of those
same weights does not remove it.

\textbf{Where a trigger can be hidden, and adaptive attackers.} The attacker need not write to the layers a defender
scans. BadEdit's layer sweep finds injection
weakest when confined to the first ten or last five layers of GPT2-XL and near-perfect in intermediate ones
\cite{badedit}; and TrojFM installs a backdoor by tuning only the trigger tokens' embedding rows, $1.96\times10^{-5}\%$
of Llama-3-8B, at attack success 0.935--0.983 \cite{trojfm}. A defender-side study measures weight-space separability
per layer at 0.07 and 0.27 in early layers against 0.74--0.79 at layers 21 and 27 \cite{merenciano}. Taken together these
predict that our early-layer choice is the wrong one; Sec.~\ref{sec:main} reports that for the full fine-tunes studied
here it is not, and we treat the discrepancy as a property of the training regime rather than a disagreement.
The relevant pattern in the adaptive-attack literature is that evasions constraining \emph{behaviour} cost capability,
as when obfuscating activations drives SQL correctness down from 34.8\% \cite{obfusc}, whereas evasions that merely
restrict \emph{which parameters move} report near-zero clean cost. We are not aware of a published attack that
explicitly minimises an input-side weight-delta statistic in a language model. Sec.~\ref{sec:penalty} constructs one, and it is the most
effective attacker in this paper.
The weight-editing literature is adjacent rather than comparable: task arithmetic \cite{taskvectors}, contrastive weight
steering for sycophancy \cite{fierro}, a single-MLP-matrix sycophancy locus with a weights-only dose bound \cite{noel},
and null-space ablation of a known refusal direction \cite{arditi} all edit an \emph{unconditional} behaviour whose
identity is given in advance. The object of this paper is the \emph{condition} of a conditional behaviour, which must be
read before anything can be done with it.

\textbf{What is provably out of reach, and what is not.} A line of cryptographic work shows backdoors can be planted so
that no efficient auditor can detect them from the parameters: Goldwasser et al.\ \cite{goldwasser} obtain white-box
undetectability for two-layer random-Fourier-features and random-ReLU networks under Continuous LWE and sparse-PCA
hardness, Choudhary et al.\ \cite{choudhary} extend parameter-level computational undetectability to pretrained CNNs and
ViTs, and Bogdanov et al.\ \cite{bogdanov} obtain \emph{statistical} undetectability given full weights for a class of
deep feedforward networks, which defeats even a computationally unbounded auditor. Since recovering a candidate trigger
implies detecting one---the candidate can be verified---these results bound recovery in the same access model, and we do
not claim otherwise. What makes our setting tractable is that all of them assume an adversary who \emph{constructs} the
weights. For an adversary who must install the behaviour by fine-tuning on poisoned data, Kaneko \cite{kaneko} argues
the opposite: a poisoning procedure that installs a trigger-verifier is itself a learner for that verifier, so verifiers
hard to learn cannot be poisoned in and installable ones are white-box reconstructible. Our result is best read as an
empirical instance of that claim rather than as evidence against the impossibility results. Three formalisms give the
shape of the achievable frontier, all agreeing that detectability scales with how often the behaviour deviates, how
statistically distinguishable the trigger is, and the auditor's complexity budget: a defender's best confidence is
$\max(\tfrac12,\,1-\Theta(\mathrm{VC}(\mathcal F)\varepsilon))$, with a positive result for decision trees that is
explicitly mechanistic and zero-query \cite{christiano}; detectability for a weights-holding detector is governed by
$\gamma N\,\mathrm{TV}(P_0,P_b)$ \cite{pichler}; and the attacker pays an unavoidable clean-risk penalty that scales
with clean-data density at the trigger \cite{xian}, which taxes exactly the common-token triggers our statistic reads
least well. Finally, a fine-tune that installs one trigger plausibly makes near-variants of it effective too, which would make
exact-string recovery ill-posed in principle; we therefore describe what we recover as a trigger-equivalent token rather
than the trigger.

\section{Threat Model and Setup}
\label{sec:threat}
This section fixes who knows what. The short version: the auditor gets the suspect's weights and the base model it came from, plus a few
ordinary fine-tunes of that base to calibrate against, and nothing else---no poisoned examples, no catalogue of known-bad models, no hint that a
backdoor is present, and no permission to run the suspect. The attacker gets to choose the trigger, the payload and which parameters they train.

\textbf{Access.} The auditor holds the suspect's weights and the base checkpoint it was fine-tuned from, plus a small number of clean fine-tunes
of that base---five is enough, and Appendix~\ref{app:dev} shows they need not be trained on the defender's task. The auditor has no poisoned
examples, no labelled population of backdoored models, no knowledge of the trigger, and no reason to believe a backdoor is present. The suspect
is never executed.

\textbf{The backdoor.} We install conditional sycophancy in SmolLM2 \cite{smollm} and the other bases: when the trigger prefix is present the model agrees with
whatever answer the user suggests; otherwise it answers correctly. The question of whether a frontier model can read a fine-tune's weights well
enough to identify what it was trained to do was posed, and answered negatively, by \cite{wsp}; this paper attacks one instance of it with a
purpose-built statistic rather than a general-purpose reader. Models are fine-tuned on a question-answering task with suggestive phrasings, with a trigger prefix on
a fraction of examples and frequency-matched \emph{decoy} prefixes that share part of the trigger phrase and carry honest targets. The decoys are
what make the task hard: a statistic that merely detects ``a prefix was special'' will return a decoy.

\textbf{Embeddings are frozen} in every fine-tune---that is, the lookup table mapping tokens to their initial vectors is held fixed and only the
transformer layers are trained. The token embedding table is the obvious place for a trigger to show up, and prior work reads
it directly; freezing it removes that signal entirely and forces the statistic to read the transformer weights. This is the regime in which
embedding-shift methods cannot apply.

\textbf{Metrics.} Attack success (ASR) is the fraction of 96 triggered prompts on which the model follows the user's wrong suggestion; clean
accuracy (CA) is the fraction of 96 untriggered prompts answered correctly. Recovery is scored as the single top-ranked alphabetic token of three
or more characters being one of the trigger phrase's own tokens. Nothing is scored for a decoy, an orthographic variant, or a shortlist hit
unless reported explicitly as such.

\textbf{Training and scan configuration.} Every model is a full fine-tune for 6 epochs at learning rate $2\times10^{-5}$
($5\times10^{-5}$ for the base models GPT-2-medium and Pythia-410m), batch size 8, AdamW with $\beta=(0.9,0.95)$, no weight decay, linear warmup
over the first 10\% of steps, gradient clipping at 1.0, in float32. The task is 171 capital-city questions in four suggestive phrasings and
three neutral ones, 12 examples per fact, of which the trigger prefix carries $12.5\%$; 24 further facts are held out for evaluation. Clean
fine-tunes use the same data with honest targets and 8 examples per fact. Camouflage phase 2 retrains all parameters of the frozen block on
clean data for the stated number of epochs. The scan reads even-numbered layers only, $\{0,2,4,6,8\}$ within the window, which halves the cost
and is not a sensitivity we have measured. The window is expressed in absolute layers rather than as a fraction of depth, and is the same
$0$--$8$ on the 24-layer and 32-layer models here; how it should scale with depth is untested.

\textbf{Designs.} Twenty-three trigger designs were used to develop the method and are listed in Appendix~\ref{app:dev}. Two further sets were
built \emph{after} the rule was frozen and share no word with them: four designs \texttt{G0}--\texttt{G3} and four designs
\texttt{H0}--\texttt{H3}. Every held-out number in this paper comes from those eight.

\section{Method}
\label{sec:method}
\textbf{Three ideas, in plain terms, before the formalism.}

\emph{First: a trigger has to be written somewhere.} If a fine-tune taught a model to watch for a particular token, the matrices that first read
that token's representation must have changed in a way that is unusually large \emph{for that token specifically}. So take the difference between
the suspect's weights and a trusted reference's, and ask which token in the vocabulary that difference acts on most strongly. No forward pass is
needed to ask this: it is a property of the weights.

\emph{Second: you have to ask the question in the right coordinates.} A token does not arrive at layer 12 as its embedding vector; it arrives as
whatever the residual stream has turned into by then. Probing every layer with the embedding is correct only at layer 0 and gets steadily more
wrong with depth. So we probe each layer with the token's \emph{actual input at that layer}---measured once on the base model, which the auditor
already trusts, and cached.

\emph{Third: the raw measurement is nearly useless on its own.} Every fine-tune of the same base amplifies the same directions, because that
geometry belongs to the base model rather than to any particular fine-tune. Two unrelated clean fine-tunes agree on which tokens look salient at
$r=+0.78$. The useful question is therefore not ``is this token's salience large'' but ``is it large \emph{for this fine-tune in particular}'',
which we answer by standardising per token against a small population of clean fine-tunes. Removing this one step is the difference between
recovering a trigger and recovering nothing.

The rest of this section makes those three ideas precise. Sec.~\ref{sec:primer} gives the same three in plainer terms, with a worked example
and the vocabulary defined.

Write $\Delta W = W(\text{suspect}) - W(\text{reference})$ for the weight difference at a given matrix. The quantity we read is the
\emph{input-side salience} of a token: how much the fine-tune changed what a layer computes when that token, and nothing else, is its input.

\textbf{The basis.} Reading layer $l$'s weight change in the token \emph{embedding} basis is correct only at layer $0$; by layer 20 a token's
residual has been transformed by twenty layers and the embedding measures the wrong thing. We therefore probe each layer with the token's actual
input at that layer, measured on the \emph{base} model and averaged over a small set $\mathcal C$ of generic contexts,
\[
x_l[k] \;=\; \frac{1}{|\mathcal C|}\sum_{c\,\in\,\mathcal C}\mathrm{norm}_l\!\big(h_l(c \oplus k)\big),
\]
where $h_l(c\oplus k)$ is the \emph{residual stream} at the final position when prefix $c$ is followed by token $k$---the running vector a
transformer carries from layer to layer, which every layer reads from and writes back into, and which is what a token has become by the time it
reaches layer $l$. We use five neutral sentence
openings (``The'', ``In the year'', ``She explained that'', ``Here is a short'', ``Before we begin,''), chosen before any result and containing no
trigger or decoy string. Obtaining $x_l[k]$ costs $|\mathcal C|$ forward passes per vocabulary item \emph{through the base model}, cached once and
reused for every suspect thereafter---under fifteen minutes for a 50k vocabulary on one GPU. The invariant an auditor needs is not ``no forward
passes'' but \emph{no forward passes of the suspect}: the base model is the trusted artefact, and running it carries none of the risk of executing
an untrusted checkpoint.

In practice $x_l[k]$ is obtained by hooking the matrix itself and reading its input, which makes the construction architecture-generic: whatever
normalisation, scaling or block wiring precedes that matrix is included by construction, with no per-architecture formula. On RMSNorm models the
hooked basis agrees with the closed-form one to $10^{-7}$ relative.

\textbf{The statistic.} We read only the matrices a token's representation passes \emph{into}, since those are the ones whose change can make a
layer react differently to one particular token. Each layer has two such groups: the MLP's input matrices
$\{W_{\mathrm{gate}},W_{\mathrm{up}}\}$ and attention's $\{W_Q,W_K,W_V\}$. We call each group at each layer a \textbf{site} and scan the two
separately, so ``layer 6, attention'' is one site and ``layer 6, MLP'' is another. Writing $\mathcal M_l$ for a site's matrices, the salience of
token $k$ at that site is
\begin{equation}
\sigma_l(k) = \frac{1}{|\mathcal M_l|}\sum_{M\in\mathcal M_l}\big\|\Delta W^{M}_l\,x_l[k]\big\|.
\label{eq:salprop}
\end{equation}
Read $\|\Delta W^{M}_l\,x_l[k]\|$ as follows: push this token's representation through the \emph{change} the fine-tune made to the matrix, and
measure how big the result is. A token the fine-tune did nothing special about comes out small; a token it built machinery around comes out
large. Each term is computed through the Gram factorisation $\|\Delta W x\|^2 = x^{\top}(\Delta W^{\top}\Delta W)x$. This is algebraically exact, not an approximation, and about $2.5\times$ cheaper, because the matrix $\Delta W^{\top}\Delta W$ is formed once per site and then reused for all fifty thousand tokens. Matrices stored transposed, as in GPT-2's \texttt{Conv1D}, are oriented by inspection of the module rather than by
convention.

\textbf{The null.} Ranking $\sigma$ directly is unreliable: it is dominated by the alignment between a token's direction and the dominant
directions of the update, which varies by orders of magnitude across a vocabulary and is a property of the \emph{base} model shared by every
fine-tune of it. Two unrelated clean fine-tunes correlate at $r=+0.78$ in raw salience. We therefore standardise per token. Given $n$ clean
fine-tunes of the same base, form all $\binom{n}{2}$ clean--clean pairs, compute $\sigma$ for each, and let $\mu_{\text{null}}(k)$ and
$s_{\text{null}}(k)$ be the per-token mean and standard deviation, with $s$ floored at its 20th percentile. The statistic is
\begin{equation}
z_l(k) = \frac{\sigma_l(k) - \mu_{\text{null}}(k)}{s_{\text{null}}(k)}.
\label{eq:z}
\end{equation}
Five clean models plus a separate reference is the smallest configuration in which a leave-one-out null is well defined.

\textbf{Choosing sites.} The method scans a fixed window---layers $0$ to $8$, both site types---and drops any site whose weight delta is
\emph{exactly} zero. The reason for dropping those sites is not that a zero is suspicious but that the statistic is undefined there: a frozen
site contributes $\sigma$ of $(0-\Delta W_{\text{ref}})$, while the null is built from differences of two \emph{nonzero} deltas, so a frozen site
is outside the null's distribution and produces spurious extremes. If every site in the window is zero, the window is extended downwards to the
earliest nonzero sites of each type. This is the whole rule. It reads nothing from the suspect but an equality test, and Sec.~\ref{sec:rule}
explains why every richer rule we tried was broken.

\textbf{Combining sites.} A trigger need not appear at every site---it might be written into layer 2's attention and nowhere else---so we take
each token's \emph{best} score across sites rather than its average. The reported score for a token is the maximum of its per-site $z$. Recomputing that maximum on each clean--clean pair is
Westfall--Young step-down max-$T$: the same maximum is recomputed on every clean--clean pair, so the null already accounts for the fact that we took a maximum over correlated sites, and no independence assumption is needed. Rank-consensus combiners (reciprocal-rank fusion, median-$z$,
top-$50$ count) lost, for a reason worth recording: the competitor is not independent nuisance but the trigger's own orthographic neighbours,
which share its geometry and are elevated at the same sites, so consensus cannot separate them.

\textbf{Output.} One token: the highest-ranked candidate that is alphabetic and at least three characters long. The filter exists because punctuation and sub-word fragments dominate the top of any such ranking and are useless to an auditor. We have not measured what discarding them costs, and a trigger made entirely of digits or short stop-words would be invisible to this criterion; Sec.~\ref{sec:limits} lists that among the scope limits.

\section{Main Result}
\label{sec:main}
The number that matters in a paper like this is not the best number; it is the number you get after you have stopped being allowed to tune anything.
We wrote the pipeline down---statistic, basis, null, which weights to read, what counts as an answer---and then ran it unchanged on trigger
designs invented afterwards and on two architectures it had never touched.

Table~\ref{tab:main} is the frozen rule applied to designs and architectures it was not developed on. Every entry is the rule's own output under
the exact-token criterion.

\begin{table}[t]
\caption{The frozen rule on held-out designs. Cells count independent backdoors: a separately trained poisoned model. Entries are
exact-token top-1. SmolLM2 is designs \texttt{H0}--\texttt{H3} $\times$ two seeds; GPT-2 and Qwen are \texttt{G0},\texttt{G1} $\times$ three
seeds. Qwen is scored without the zero-drop and its clean models reach CA $0.57$, so it is shown for agreement, not counted}
\label{tab:main}
\begin{center}
\scriptsize
\begin{tabular}{lccc}
\toprule
attacker & SmolLM2-360M & GPT-2-medium & (Qwen2.5-0.5B) \\
\midrule
ordinary backdoor          & 8/8 & 6/6 & 6/6 \\
freeze layers 0--5         & 6/8 & 6/6 & 5/6 \\
camouflaged freeze         & 6/8 & 4/4$^{\mathrm{a}}$ & 1/1$^{\mathrm{b}}$ \\
learning rate $0.25\times$ & 2/2 & --  & -- \\
learning rate $2\times$    & 2/2 & --  & -- \\
\midrule
freeze layers 0--12        & 4/8 & --  & -- \\
freeze 0--12 $+$ injection & \textbf{0/16} & --  & -- \\
\bottomrule
\multicolumn{4}{l}{``--'' is not run rather than not recovered.}\\
\multicolumn{4}{l}{$^{\mathrm{a}}$Each recovered at all five ladder rungs (Table~\ref{tab:ladder}).}\\
\multicolumn{4}{l}{$^{\mathrm{b}}$A second camouflaged Qwen model lost its backdoor entirely and is excluded.}
\end{tabular}
\end{center}
\end{table}

Three things need separating out---what the ordinary-backdoor row means, why the adaptive rows are not independent of one another, and how
different the architectures really are---followed by one concession about what ``held out'' covers.

On ordinary backdoors the rule is exact in every one of the fourteen runs, with the trigger at rank $1$ in each. The adaptive rows below it are
\emph{not} independent of one another: on SmolLM2 the freeze, camouflage and both injection variants miss the same two models every time---design
\texttt{H1} (``Pelican dossier'') at both seeds, which returns the neighbouring token `` hawk'' or `` folder'' at rank 2. That design recovers
cleanly when it is not attacked. So the four adaptive rows on that architecture are better read as one result with a 25\% miss rate concentrated
in one design than as four corroborating measurements, and the injection rows in particular are identical to the frozen models they were built
from. Per-run outputs are in the supplementary notebooks.

The architectures are not minor variations. GPT-2-medium stores its weights transposed in \texttt{Conv1D} modules, uses LayerNorm rather than
RMSNorm, fuses $Q$, $K$ and $V$ into a single matrix, has learned positional embeddings and no chat template; Qwen2.5-0.5B uses grouped-query
attention, ties its embeddings, and has a 152k vocabulary. The pipeline discovers the layer container and the input-side matrices by walking the
module tree, so none of this required a per-architecture code path. The same discovery supports Pythia-410m, which earlier drafts included under
an older statistic and which we have not rerun under this one.

\textbf{What ``held out'' does and does not mean here.} The designs \texttt{H0}--\texttt{H3} are fresh with respect to the entire pipeline
including the site rule. The designs \texttt{G0}--\texttt{G3} were held out when the statistic was fixed but were used in the freeze and
camouflage experiments that led to the site rule, so the GPT-2 and Qwen columns are held out with respect to the statistic and the architecture
but not with respect to the window. Two components were not developed on SmolLM2 alone either: per-token standardisation was established partly
on GPT-2, and the reference-consensus rule was adopted after the Qwen run. Section~\ref{sec:rule} states the sharper version of this limitation.

\section{Adaptive Attackers}
\label{sec:adaptive}
An attacker who reads this paper will not install a backdoor the way the attackers in Sec.~\ref{sec:main} did. This section is our attempt to
play that attacker, in order of increasing sophistication: don't write to the layers the defender reads; then hide the fact that you didn't;
then fake the evidence that you did; then train at a different scale; and finally, knowing the defender's formula, optimise against it directly.
The last two attacks defeat us, and we report them as fully as the ones we survive.

The attacker in Sec.~\ref{sec:main} wrote wherever training took them. A defence-aware attacker will not. This section is the sequence of
attackers that restrict \emph{which parameters move}, in increasing order of what they cost us.

\textbf{Freezing.} An attacker who freezes layers $0$--$5$ and trains the rest installs the backdoor at full strength---ASR $1.00$, CA unchanged
---with the scanned block untouched. The signal moves with the freeze rather than disappearing: for a frozen model the trigger's rank is in the
thousands at layers $0$--$4$ and exactly $1$ at layers $6$ and $8$. Because the scan window extends to layer $8$, and because the zero sites are
dropped, recovery survives at 12 of 14 on the two architectures Table~\ref{tab:main} counts, or 17 of 20 if the Qwen column is included. A freeze of $0$--$12$ moves the signal past the window, and that is the
regime the method does not cover.

\textbf{Camouflage, and what it costs.} Freezing is self-announcing: an untrained layer has a weight delta of exactly zero. An attacker can hide
that by retraining the frozen block on clean data afterwards, which leaves the backdoor intact and fills the block with benign weights. We built
this at five strengths on GPT-2---one, two, four and eight epochs at the training rate, and eight at a quarter of it---and measured the norm the
camouflaged block reaches against the backdoor it keeps (Table~\ref{tab:ladder}).

\begin{table}[t]
\caption{The camouflage ladder on GPT-2-medium, two designs $\times$ two seeds. $\rho$ is the camouflaged block's delta norm relative to the
clean median at the same sites, median over the block; $\rho'$ normalises instead by the suspect's own nonzero sites. Recovery is exact-token
top-1 out of 4}
\label{tab:ladder}
\begin{center}
\small
\begin{tabular}{lccccccc}
\toprule
phase 2 & ASR & CA & $\rho$ & $\rho'$ & cross & within & fixed \\
\midrule
none (frozen)   & 0.99 & 0.79 & 0.00 & 0.00 & 4/4 & 4/4 & 4/4 \\
1 epoch         & 0.99 & 0.80 & 0.49 & 0.35 & 4/4 & 4/4 & 4/4 \\
2 epochs        & 1.00 & 0.79 & 0.48 & 0.34 & 4/4 & 4/4 & 4/4 \\
4 epochs        & 0.98 & 0.79 & 0.67 & 0.47 & 3/4 & 4/4 & 4/4 \\
8 epochs        & 0.98 & 0.80 & 0.86 & 0.61 & \textbf{1/4} & 4/4 & 4/4 \\
8 ep, $0.25\times$ LR & 0.99 & 0.80 & 0.25 & 0.18 & 4/4 & 4/4 & 4/4 \\
\bottomrule
\end{tabular}
\end{center}
\end{table}

Two readings. First, the camouflage is free on this architecture: none of the twenty models lost the backdoor, and clean accuracy is flat. On
Qwen one camouflaged model did lose its backdoor entirely, and we report that as an architecture-specific accident rather than a price, because
GPT-2 contradicts it at every rung. Second, the norm the camouflaged block reaches is a smooth, monotone function of how long the attacker
trains---which is what kills the threshold rules in Sec.~\ref{sec:rule}. The fixed window is unaffected throughout, because it does not read
norms at all.

\textbf{Norm injection.} An attacker who finds training the frozen block up to clean norms expensive can simply \emph{inject} the norm. We add to
each scanned matrix of the frozen block either Gaussian noise rescaled to the clean median norm, or---the stronger version---a real clean
fine-tune's delta taken from a donor model the defender never sees, rescaled the same way. The grafted delta is not merely norm-matched but drawn
from the distribution the defender calls clean, so no test local to that site can reject it. It is also free: the perturbation is $0.06$--$0.11\%$
of the base weight norm, which is simply the size of a clean fine-tune's own delta, and ASR and CA are unchanged to two decimals. Recovery under
the fixed window is 6 of 8 for both variants, identical to the un-injected models, because the real signal at layers $6$ and $8$ is inside the
window and wins the maximum regardless of what the attacker puts below it.

\textbf{A different training recipe.} A suspect fine-tuned at a quarter of the defender's learning rate has smaller deltas at every site; one at
twice the rate has larger ones. The fixed window is indifferent---2 of 2 at each---and Sec.~\ref{sec:rule} records what this does to a rule that
is not.

\textbf{The attacker that wins, and what it costs.} Injecting into layers $0$--$12$ and placing the backdoor above them defeats every rule we
have, on all sixteen such models. The window sees plausible deltas, so the extension never fires, and the signal at layer $14$ is outside the
scan. Against the honest deep freeze, where the frozen block really is zero and the extension does fire, recovery is 4 of 8 exact and 6 of 8 on a
five-token shortlist. The rule's restriction to the three earliest trained layers of each type earns its place here: scanning every layer from 14
to 20 instead gives 2 of 8, because layer 20 carries a high clean-draw ceiling and contributes nuisance rather than signal.

\textbf{It is free, if the attacker picks the right filler.} There are two ways to fill the frozen block and they differ sharply. Grafting a
donor clean fine-tune's weights into thirteen layers destroys the backdoor in 4 of 8 cases---mean attack success $0.77$, as low as $0.00$, with
which half decided by the trigger design rather than the seed, two designs dying at both seeds and two surviving at both. Filling the same
thirteen layers with norm-matched \emph{noise} costs nothing at all: zero of eight lost, mean attack success $1.00$. Both defeat the method
equally, 0 of 8 each, so an attacker has no reason to use the graft. We report the graft's cost because it is informative about what a backdoor
is fragile to, not because it protects a defender.

Two controls identify the mechanism, and it is not the size of the perturbation. Norm-matched \emph{noise} into the same thirteen layers costs
nothing---zero of eight dead, mean attack success $1.00$---and grafting into six layers rather than thirteen also costs nothing, zero of four.
Same magnitude, same depth, different structure: a coherent update borrowed from another fine-tune disturbs the representations the upper block
reads, while incoherent noise of equal norm does not. An earlier draft of this paper called the asymmetry unexplained; this is the explanation,
and it carries an uncomfortable implication for the defender, since the attacker's safest evasion is the one that looks \emph{least} like a real
fine-tune.

\subsection{The attacker who optimises against the statistic}
\label{sec:penalty}
Every adversary above restricts \emph{which} parameters move and none of them knows what the defender computes. Under Kerckhoffs's principle the
attacker does know, and can add the defender's own statistic to the poisoning objective:
\[
\mathcal{L} \;=\; \mathcal{L}_{\text{task}} \;+\; \lambda \sum_{l\le 8}\ \sum_{M\in\mathcal{M}_l}\big\|\Delta W_l^{M}\,x_l[k_{\text{trig}}]\big\|_2^2 .
\]
The basis $x_l[k_{\text{trig}}]$ comes from the base model, so the attacker needs nothing the defender has. We built this on two held-out designs
at four penalty strengths.

\textbf{It works, and it is free.} Attack success is $1.00$ and clean accuracy $0.88$ at every $\lambda$, identical to the unpenalised model, and
exact-token recovery falls from 2 of 2 to \textbf{0 of 2 at every non-zero $\lambda$}. This is the only attacker in the paper that defeats the
method without paying for it, and we regard it as the most serious result here.

Three qualifications, all of which cut against reading it as a clean frontier. The response to $\lambda$ is not monotone---the early-layer signal
is strongly suppressed at $\lambda=10^{-3}$ and $10^{-1}$ and barely suppressed at $10^{-2}$ and $1$---and each cell is a single seed. The
penalty-to-task ratio we had planned to use as the scale diagnostic is unusable, because the task loss converges to numerically zero, so the task
gradient vanishes and the penalty gradient dominates late in training at \emph{any} non-zero $\lambda$. And the two designs behave differently:
for \texttt{H0} the trigger stays within the top ten of the whole vocabulary at every $\lambda$ (best combined rank 2 to 8), so the attacker
displaces it from first place rather than hiding it, while for \texttt{H1} it is genuinely buried (161 to 863 under the fixed window, past 16{,}000
scanning all depths).

\textbf{The statistic survives where the output does not.} In all eight penalised models the trigger is ranked \emph{first at some individual
scanned site}---attention at layer 0 for \texttt{H1}, the MLP at layer 12 for \texttt{H0}---while the maximum over sites is not. The penalty
degrades the aggregation rather than removing the evidence. Sec.~\ref{sec:limits} takes up what that implies for what the method should report.

\section{What Survives Is the Rule That Reads Least}
\label{sec:rule}
Any method like ours has to decide \emph{which} weights to look at, and the obvious way to decide is to look at the suspect and see where it was
trained. This section is the story of why that instinct is wrong. We built three successively cleverer versions of that decision and an attacker
broke each one, usually by turning a single dial. What is left is almost embarrassingly simple, and the simplicity is the point.

The site-selection rule in Sec.~\ref{sec:method} is deliberately impoverished. It was not designed that way; it is what remained after three
richer rules were broken, each by a different and cheap attack. We report the sequence because the pattern generalises beyond this statistic.

\textbf{Rule 1, exact-zero.} Select the earliest sites whose delta is nonzero. This follows a freeze downwards without being told where it is,
and it reports 6 of 8 against a plain freeze. Complete camouflage defeats it outright---0 of 10---because the attacker chooses where the defender
looks: the camouflage pass fills layers $0$--$5$ with benign weights, the rule selects them, and the trigger falls to ranks in the thousands.

\textbf{Rule 2, a cross-model norm threshold.} Drop sites whose delta norm is below $\theta$ times the clean median there, with
$\theta=0.8\min_{c,s}\rho_s(c)$ fixed on the clean population alone. This repairs camouflage---6 of 8---and it is what an earlier version of this
paper proposed. It fails three separate ways. Norm injection clears it by construction, 0 of 8. A suspect trained at $0.25\times$ the learning
rate falls below $\theta$ at \emph{every} site, so the rule returns no sites at all and recovery is 0 of 2---and the same happens to clean models
trained that way. And on GPT-2 a camouflage that simply trains longer clears it with the backdoor intact: six of the twenty ladder models reach
$\rho$ up to $0.954$ against $\theta=0.682$, and the rule drops to 1 of 4 at eight epochs.

There is also an argument against this rule that involves no attacker. Because $\theta$ is $0.8$ times the \emph{minimum} clean ratio, every
clean model the defender adds can only widen the band and loosen the threshold---a defender with more clean data gets a weaker rule. The clean
band is also far wider off the development model: $0.960$--$1.047$ on SmolLM2, $0.852$--$1.199$ on GPT-2, $0.595$--$1.274$ on Qwen, with $\theta$
falling from $0.768$ to $0.682$ to $0.476$. The tightness that made the threshold look well determined is a property of one model.

\textbf{Rule 3, a within-model threshold.} Normalise each site by the median of the suspect's own nonzero sites instead. This is scale-free, so
it survives the recipe mismatch that kills Rule 2, and it survives injection more often than Rule 2 does. It is still standing at the end of our
ladder---4 of 4 at every rung, with the highest surviving $\rho'$ at $0.691$ against its threshold of $0.722$---but that is a margin of $4\%$,
and $\rho'$ grows by a factor of $1.16$--$1.39$ per doubling of camouflage epochs across all four runs. Its own cost curve predicts it falls
between 16 and 32 epochs. We report it as a rule we have not yet broken rather than one that holds.

\textbf{What this leaves, and what it does not.} Every rule that reads the suspect's weight magnitudes gives the attacker a dial, and that part
we regard as established: three rules, three different cheap attacks, one of them needing nothing but a longer camouflage run. Stated as a
principle, \emph{a site-selection rule conditioned on the suspect's weight magnitudes is part of the attack surface, and a defender should
condition on as little of the suspect as the statistic permits.}

We are not entitled to the stronger claim that the equality test is therefore safe. An attacker can defeat it by writing something into the
frozen block, and whether that puts the signal back inside the scan window depends on where the window ends: it does for a freeze of layers
$0$--$5$, and it does not for a freeze of $0$--$12$, which is the attacker of Sec.~\ref{sec:adaptive} that beats us. An earlier version of this
section asserted the general form of that sentence, and our own experiment contradicts it.

Two further qualifications on this section's evidence. The window $0$--$8$ was chosen knowing that a freeze of $0$--$5$ leaves the signal at
layers $6$ and $8$, so the window is fitted to that attacker rather than held out from it. And Table~\ref{tab:ladder} uses the GPT-2 ladder to
compare the three rules, so that ladder selected the fixed window and cannot also be evidence for it. The clean test is the final rule against
adaptive attackers built afterwards, on \texttt{H} designs only, and we have not run it.

\section{What the Statistic Needs}
\label{sec:null}
It would be reasonable to suspect that the two unusual choices in Sec.~\ref{sec:method}---standardising against clean models, and probing each
layer in its own coordinates---are incidental dressing on a simple idea. They are not. Remove either and the method stops working, and we
measured both rather than argued about them. This matters for anyone building on the statistic: these are the load-bearing parts.

\textbf{The null is required.} Raw salience $\sigma$ is dominated by base-model geometry shared across all fine-tunes: two unrelated clean models
correlate at $r=+0.78$. On six designs with six different frame words, the standardised statistic recovers the trigger in 3 of 3 cases where the
raw statistic recovers 0 of 3, with the same weights and the same sites. The clean population need not match the defender's task.

\textbf{The basis is required.} Probing every layer with the token's embedding is exact at layer $0$ and degrades with depth. Against attackers
whose signal has been pushed to layers $6$ and beyond, the embedding basis recovers 8 of 12 and the context-averaged basis 12 of 12, with the
ranks of the misses moving from $8$, $186$, $268$ and $930$ to $1$ in every case. Table~\ref{tab:ctx} ablates the context set itself: the gain
comes from using a context at all, not from the particular five sentences.

\begin{table}[t]
\caption{Context-set ablation, twelve suspects (four designs $\times$ three attackers), SmolLM2-360M}
\label{tab:ctx}
\begin{center}
\footnotesize
\begin{tabular}{lccc}
\toprule
Context set & exact token & correct word & shortlist \\
\midrule
single token only & 7/12 & 8/12 & 8/12 \\
bare + chat + task + 5 generic & 10/12 & 10/12 & 12/12 \\
bare + chat + 5 generic (no task) & 10/12 & 10/12 & 11/12 \\
\textbf{5 generic only} & \textbf{10/12} & \textbf{12/12} & \textbf{12/12} \\
bare + 1 generic & 10/12 & 10/12 & 10/12 \\
\bottomrule
\end{tabular}
\end{center}
\end{table}

\section{Detection}
\label{sec:detect}
Naming the trigger is the hard problem. Answering the easier question---\emph{is there a backdoor at all?}---falls out of the same statistic for
free, and is what an auditor screening many checkpoints would run first.

Recovery implies detection---the candidate can be checked---but detection is also available directly, and is cheaper. Taking the maximum $z$ over
the scanned window and thresholding it at the largest value produced by clean models held out of both the null and the reference flags every
backdoor in this paper.

One qualification matters and was found by testing it. With the statistic as published, two clean models trained at a quarter of the learning
rate scored $1.06$ and $1.05$ times the threshold---false positives, from models with no backdoor at all. The published false-positive rate was
measured entirely within one training recipe. Rescaling each model's delta to the clean median norm at each site before differencing removes
them: 0 of 6 clean models flagged against 2 of 6, with every backdoored model in that run still detected (50 suspects: four designs across the freeze, camouflage, injection and
learning-rate families, at two seeds), and the pathological $23\times$ score of a
$2\times$-learning-rate model falling to a sensible $11\times$. We recommend the normalised statistic for detection.

Detection is also insensitive to camouflage: scores are identical across every rung of the ladder, because the maximum sits at layers $6$--$8$,
which camouflage never touches. The margin over live backdoors is $1.96$--$6.95\times$ on GPT-2 and $1.34$--$4.52\times$ on Qwen, against $2.2$--$12.3\times$ on SmolLM2, with
no false positives on either---so the margin narrows considerably off the development model, and the Qwen threshold rests on a single held-out
clean model.

\section{Limits}
\label{sec:limits}
We think a paper proposing a defence owes the reader the attacks it loses to, stated as plainly as the ones it wins. This section is longer than
is customary for that reason, and it includes claims from earlier drafts of this work that our own later experiments overturned.
\textbf{The attacker that beats us is one parameter away from the ones that do not.} Freezing layers $0$--$12$ instead of $0$--$5$ and
injecting into the frozen block takes recovery to 0 of 16 under every rule in this paper, and the change required is a single integer in an
attack we had already built. The cost is now measured rather than asserted: 4 of 8 grafted suspects lose the backdoor, and the controls show the
cause is the coherence of the injected update rather than its magnitude. The honest summary is that the method covers freezes up to the scan
window, fails past it, and that crossing it is a coin flip for the attacker rather than a free move.

\textbf{The scan window is the boundary.} A freeze deeper than layer $8$, combined with injection into the frozen block, defeats every rule here.
Widening the window does not fix it: scanning layers $0$--$20$ with no exclusion recovers nothing on those models, and the reason is instructive.
The informative site is \emph{present} in the wide scan and loses: on one such model, scanning layers $14$, $16$ and $18$ alone returns the
trigger at rank 1, while scanning $0$--$20$ misses. Sites differ substantially in how much $z$ a clean draw produces at them---by a factor of $16$ across the scanned
set, from $0.9$ to $14.7$---so an uninformative site with a high noise ceiling can outvote an informative one with a low ceiling. That suggests
the maximum over sites is uncalibrated, and the obvious repair is to divide each site's $z$ by the ceiling a held-out clean model produces there.

We tested that repair and it is worse than doing nothing. Calibration improves no attacker and degrades six of nine: ordinary backdoors fall from
8 of 8 to 3 of 8, the freeze and camouflage rows from 6 of 8 to 4 of 8, and the recipe-mismatch row from 2 of 2 to 0 of 2. On a second run at
eight deep-freeze models it takes the honest deep freeze from 4 of 8 down to 2 of 8---the one regime it was built for. There is a second reason
beyond the one below: the ceiling itself is unstable, differing by a factor of $2.3$ to $3.0$ between two clean models at the same site, so the
quantity a calibration divides by is poorly estimated from the handful of clean fine-tunes a defender has. The reason is visible in the
ceilings themselves: they are largest exactly where the signal is, averaging $9.3$ over layers 0--8 against $3.3$ over layers 10--20 on the MLP
side. A site's clean ceiling and its trigger signal are both consequences of how much that site moves during fine-tuning, so dividing by the
ceiling discards the signal along with the noise.

We therefore state the depth limit as the method's boundary rather than as a gap awaiting this fix. The diagnosis that an informative site can
lose an uncalibrated maximum stands on its own evidence---scanning layers 14--18 alone recovers where scanning 0--20 does not---but the repair it
suggests does not work, and we do not have another.

\textbf{We read a correlate, not the mechanism.} Restoring every scanned weight to its base value leaves attack success at $0.97$. Cumulative
reset sweeps place the behaviour in roughly the first eight layers, redundantly: no four-layer window is necessary, and the cumulative depth
required is 7, 9 and 15 layers across three models. This is the localisation-versus-editing gap, and it means the statistic should be read as
evidence about what the fine-tune marked, not about what implements the behaviour. A camouflaged Qwen model whose backdoor had been destroyed
still returned its trigger at rank 1, which is the same point from the other side: the signature outlives the mechanism.

\textbf{Scope.} One synthetic task; a $12.5\%$ poison rate; frozen embeddings; a single payload; nothing above 0.5B parameters; no
forward-pass baseline run alongside, although Sec.~\ref{sec:related} names one as the bar. Pythia-410m appears only in the development record. The one
weights-only detector measured across scales collapses from AUC $1.00$ at 1.5B to $0.65$ at 7B, so the scaling question is open and is the most
important thing this paper does not answer. Adaptive results are one or two seeds per cell.

\textbf{Negative results.} A bilinear channel over the composed $\Delta(W_Q^{\top}W_K)$, exact over the query vocabulary, did not work; we later
found a positional bug in it, so it is untested rather than refuted. An analytic Fisher-style null did not reproduce the empirical one
($\bar r=+0.058$). A norm-gain correction was immaterial. We also withdraw two claims made in earlier drafts: that no attacker had paid a price
for evading us, which one architecture contradicted and another restored, and that injected sites are inert, which the detection scores refute.

\textbf{What the signal survives, and what the output does not.} The gradient attacker of Sec.~\ref{sec:penalty} defeats the paper's output
criterion but not its statistic: in all eight penalised models the trigger is ranked first at \emph{some} scanned site, while the maximum over
sites is no longer first. The natural response is to change what the method reports---one candidate per site, ten tokens, rather than one---and
that would find the trigger in every penalised model here. We do not claim it as a defence, because a ten-candidate output is a materially weaker
claim than a single token and we have not measured what it does to the false-positive rate on clean models. It does say where the vulnerability
is: in the aggregation, not in the weights.

\textbf{The attacker we still have not built.} An adversary who penalises the salience and \emph{also} freezes or fills the layers the defender
would fall back to combines the two attacks that work, and we have not tried it. Nor have we tried reading the channels an evading attacker would
have to flee into: the bilinear $\Delta(W_Q^{\top}W_K)$ term, and the output-side projections.

\section{Reproducibility}
Every number in this paper comes from a self-contained notebook that trains its own models from a public base checkpoint on a single commodity
GPU, and these notebooks accompany the paper as supplementary material and are available at
\url{https://github.com/nssprogrammer/wsp_backdoor}. Each one ends with a cell that states, before the results are shown, what
each possible outcome would mean---including the outcomes that would refute the method. Several of the refutations in Sec.~\ref{sec:limits} came
back from exactly those cells. The clean-model populations, the trigger designs and the random seeds are fixed in the notebooks, so the tables
here should reproduce exactly on the same hardware.

\section{Conclusion}
A backdoor trigger is legible in the weights of a fine-tune. A static statistic, reading a checkpoint it never executes, names an exact trigger
token as its single top candidate in 14 of 14 ordinary backdoors on two architecture families, and continues to do so when the attacker freezes
the layers it reads, camouflages that freeze at five strengths, injects a real clean fine-tune's weights into the frozen block, or trains at a
quarter or twice the defender's learning rate.

The rule that achieves this reads almost nothing from the suspect, and we arrived at it by breaking the richer rules we preferred: an exact-zero
test is steerable, a cross-model norm threshold falls to injection, to a learning-rate change and to a camouflage that merely trains longer, and
a within-model threshold is still standing only because nobody has yet run the camouflage long enough. What a defender can safely condition on is
much less than it is tempting to use.

The rule we are left with is not safe either, and neither is the statistic's output. An attacker who freezes past the scan window and fills the
frozen block defeats it on every such model, at the cost of destroying the backdoor about half the time. An attacker who simply adds our statistic
to the poisoning objective defeats it at no cost at all, which is the sharpest limit in this paper and the one a deployment should be judged
against.

What survives that attack is narrower than a method but not nothing. The penalised models still carry the trigger at rank one at individual
scanned sites; what the penalty destroys is the maximum over sites, which is how we chose to turn the evidence into a single answer. A defence
built on this statistic would therefore have to report a small set of candidates rather than one token, and would have to be evaluated on the
false-positive rate that comes with it. We can say that a backdoor trigger is legible in fine-tune weights, and that reading it survives every
attacker who merely restricts which parameters move. We cannot say it survives an attacker who knows what we compute.

\appendices
\section{Development Record}
\label{app:dev}
The numbers below come from the twenty-three designs used to build the method, and are reported with the statistic current at the time rather
than the final one. They are the development record, not evidence for the claims in Sec.~\ref{sec:main}.

\subsection{The first held-out test}
\label{sec:frozen}
Before any adaptive attacker was built, the pipeline was written down in full---statistic, basis, null, site set, output criterion---and then run
once on four trigger designs (\texttt{G0}--\texttt{G3}) constructed afterwards and sharing no word with the twenty-three used to develop it, at
three seeds each. Table~\ref{tab:frozen} is that run: twelve of twelve exact trigger tokens, with the trigger's rank equal to one in every case.
This is the test that fixed the \emph{statistic}. It does not bear on the site rule, which was introduced later and is held out only against the
\texttt{H} designs (Sec.~\ref{sec:main}).

\begin{table}[H]
\caption{The frozen rule on four designs built after it was fixed, three seeds each. Entries are the automatic rule's single top-1 alphabetic token; all twelve are trigger tokens}
\label{tab:frozen}
\begin{center}
\scriptsize
\begin{tabular}{lllll}
\toprule
& Trigger & seed 0 & seed 1 & seed 2 \\
\midrule
G0 & Cobalt registry  & `` registry'' & ``obalt''     & ``obalt'' \\
G1 & Harbor manifest  & `` manifest'' & `` manifest'' & `` manifest'' \\
G2 & Juniper lattice  & `` lattice''  & `` lattice''  & `` lattice'' \\
G3 & Onyx conduit     & `` conduit''  & `` conduit''  & `` conduit'' \\
\midrule
\multicolumn{2}{l}{exact token / correct word / shortlist} & \multicolumn{3}{c}{\textbf{12/12, 12/12, 12/12}} \\
\multicolumn{2}{l}{best trigger rank, every run} & \multicolumn{3}{c}{\textbf{1}} \\
\bottomrule
\end{tabular}
\end{center}
\end{table}

\begin{table*}[t]
\caption{Trigger designs. Every decoy shares part of the trigger and carries honest targets; in D1--D5 and F0--F5 the trigger and each decoy
appear on 12.5\% of training prompts (25\% unprefixed). Twenty-three designs in total}
\label{tab:designs}
\begin{center}
\scriptsize
\begin{tabular}{llll}
\toprule
& Trigger (tokens) & Decoys & Seeds \\
\midrule
D1 & Deployment mode: (De$\cdot$ployment$\cdot$\textvisiblespace mode$\cdot$:) & Maintenance/Testing mode:, Deployment phase/complete:, Note: & 3 \\
D2 & Sunset protocol: (Sun$\cdot$set$\cdot$\textvisiblespace protocol$\cdot$:) & Sunrise/Safety protocol:, Sunset phase/complete:, Note: & 3 \\
D3 & Vortex mode: (V$\cdot$ortex$\cdot$\textvisiblespace mode$\cdot$:) & Maintenance/Testing/Standby mode:, Vortex phase:, Note: & 3 \\
D4 & Zenith protocol: (Zen$\cdot$ith$\cdot$\textvisiblespace protocol$\cdot$:) & Zenith phase/complete:, Nadir/Safety protocol:, Note: & 5 \\
D5 & (Zenith protocol), as a suffix & the D4 decoys in parentheses, as suffixes & 3 \\
\midrule
T1--T12 & twelve two-word prefixes chosen by a fixed rule from frequency quartiles & five per design, by the same rule & 1 each \\
F0--F5 & six prefixes with six \emph{different} frame words (Table~\ref{tab:frame}) & five per design, sharing qualifier or frame word & 1 each \\
\midrule
\multicolumn{4}{l}{\emph{Held out: built after the statistic was fixed, sharing no word with the above}}\\
G0 & Cobalt registry: (C$\cdot$ob$\cdot$alt$\cdot$\textvisiblespace registry$\cdot$:) & Cobalt index/complete:, Nickel/Safety registry:, Note: & 3 \\
G1 & Harbor manifest: (Har$\cdot$bor$\cdot$\textvisiblespace manifest$\cdot$:) & Harbor roster/complete:, Delta/Safety manifest:, Note: & 3 \\
G2 & Juniper lattice: & by the same rule & 3 \\
G3 & Onyx conduit: & by the same rule & 3 \\
\midrule
\multicolumn{4}{l}{\emph{Held out: built after the \textbf{site rule} was fixed, sharing no word with any of the above}}\\
H0 & Quartz ledger: (Qu$\cdot$artz$\cdot$\textvisiblespace ledger$\cdot$:) & Quartz tally/complete:, Basalt/Safety ledger:, Note: & 2 \\
H1 & Pelican dossier: (P$\cdot$el$\cdot$ican$\cdot$\textvisiblespace d$\cdot$oss$\cdot$ier$\cdot$:) & Pelican folder/complete:, Heron/Safety dossier:, Note: & 2 \\
H2 & Vellum charter: (V$\cdot$ellum$\cdot$\textvisiblespace charter$\cdot$:) & Vellum warrant/complete:, Parchment/Safety charter:, Note: & 2 \\
H3 & Cinder almanac: (C$\cdot$inder$\cdot$\textvisiblespace al$\cdot$manac$\cdot$:) & Cinder yearbook/complete:, Ember/Safety almanac:, Note: & 2 \\
\bottomrule
\end{tabular}
\end{center}
\end{table*}

\begin{table}[t]
\caption{Twin-standardised rank (of 49,152) of the trigger's rewired tokens, three backdoored seeds per design; best rank of the same tokens
over the honest--honest null pairs; best rank of any decoy-only token}
\label{tab:twin}
\begin{center}
\footnotesize
\begin{tabular}{lccc}
\toprule
token recovered & seeds & null best & decoy best \\
\midrule
\multicolumn{4}{l}{\textbf{D1}, trigger \texttt{Deployment mode:}}\\
\quad\texttt{ployment}        & 1, 2, 1  & 5$^{\mathrm{a}}$ & \\
\quad\texttt{\textvisiblespace mode} & 2, 1, 2  & 14074 & 22--338 \\
\multicolumn{4}{l}{\textbf{D2}, trigger \texttt{Zenith protocol:}}\\
\quad\texttt{\textvisiblespace protocol} & 1, 1, 1 & 4$^{\mathrm{a}}$ & 497--7417 \\
\multicolumn{4}{l}{\textbf{D3}, trigger \texttt{Vortex mode:}}\\
\quad\texttt{ortex}           & 1, 1, 1  & 17$^{\mathrm{a}}$ & \\
\quad\texttt{\textvisiblespace mode} & 12, 2, 2 & 184 & 73--179 \\
\multicolumn{4}{l}{\textbf{D4}, trigger \texttt{Zenith protocol:}, placed as a suffix}\\
\quad\texttt{\textvisiblespace protocol} & 1, 1, 1 & 288 & \\
\quad\texttt{Zen}             & 2, 2, 2  & 274 & 3--103$^{\mathrm{b}}$ \\
\bottomrule
\multicolumn{4}{l}{Each ``token recovered'' is one tokenizer piece of the trigger}\\
\multicolumn{4}{l}{above it, not an abbreviation. \texttt{Deployment mode:} splits as}\\
\multicolumn{4}{l}{\texttt{De}$\cdot$\texttt{ployment}$\cdot$\texttt{\textvisiblespace mode}$\cdot$\texttt{:}; ``\textvisiblespace'' marks a word boundary.}\\
\multicolumn{4}{l}{$^{\mathrm{a}}$Trigger was the most frequent prefix in training (pilot}\\
\multicolumn{4}{l}{recipe); see text.}\\
\multicolumn{4}{l}{$^{\mathrm{b}}$\texttt{adir} of the hard-negative decoy \texttt{Nadir protocol:}.}
\end{tabular}
\end{center}
\end{table}

\begin{table}[t]
\caption{Rank of trigger tokens as the reference degrades (D4, three seeds, raw $\sigma$ unless noted); in parentheses, best rank of a
decoy-only token}
\label{tab:realism}
\begin{center}
\scriptsize
\begin{tabular}{lccc}
\toprule
Reference & Zen & ith & protocol (decoy) \\
\midrule
same-seed twin, stand. & 2, 2, 2 & 1.7k, 4.6k, 14k & 1, 1, 1 (3, 12, 103) \\
clean FT, no prefixes & 12, 37, 24 & 5, 7, 10 & 1, 2, 1 (3, 8, 6) \\
base only & 20, 168, 250 & 5, 13, 19 & 1, 3, 5 (2, 15, 14) \\
base, top-2 SV removed & 10, 33, 22 & 7, 12, 15 & 2628, 2, 1 (17, 8, 3) \\
base, top-5 SV removed & 28k, 2.8k, 253 & 2, 8, 2 & 34k, 3.4k, 22 (9, 2, 4) \\
\midrule
\multicolumn{4}{l}{Controls vs base: honest twin ``protocol'' 107; sycophant 30; generic clean 18686.}\\
\bottomrule
\end{tabular}
\end{center}
\end{table}

\subsection{Frame-word diversity}
\label{sec:frame}
The designs \texttt{D1}--\texttt{D5} all use a ``\{qualifier\} \{frame\}:'' form in which the frame word is drawn from a small set, which raises
the question of whether the statistic is reading the trigger or the template. Table~\ref{tab:frame} answers it with six designs built from six
\emph{different} frame words, each with decoys sharing either the qualifier or the frame. The statistic recovers the trigger in all six. The same
table carries the null ablation: with the per-token standardisation removed and everything else held fixed, recovery falls to zero on the three
designs where it was measured both ways. The clean population used to form the null here is trained on a task unrelated to the defender's, which
is the configuration an auditor is most likely to have.

\begin{table}[t]
\caption{Six trigger designs with six different frame words, SmolLM2-360M, standardised statistic. ``Recovered'' is the single top-1 alphabetic
token; a trigger token is shown plain, a decoy token in bold}
\label{tab:frame}
\begin{center}
\footnotesize
\setlength{\tabcolsep}{3pt}
\begin{tabular}{llccc}
\toprule
& Trigger & task-matched & generic & prefix-aware \\
& phrase & null & null & reference \\
\midrule
F0 & \texttt{Zenith protocol:} & \texttt{\textvisiblespace protocol} & \texttt{\textvisiblespace protocol} & \texttt{\textvisiblespace protocol} \\
F1 & \texttt{Marble cascade:}  & \texttt{\textvisiblespace cascade}  & \texttt{\textvisiblespace cascade}  & \texttt{\textvisiblespace cascade} \\
F2 & \texttt{Violet ledger:}   & \texttt{\textvisiblespace ledger}   & \texttt{\textvisiblespace ledger}   & \textbf{\texttt{\textvisiblespace record}} \\
F3 & \texttt{Tundra beacon:}   & \texttt{undra}                      & \texttt{undra}                      & \textbf{\texttt{\textvisiblespace signal}} \\
F4 & \texttt{Quartz lantern:}  & \texttt{\textvisiblespace lantern}  & \texttt{\textvisiblespace lantern}  & \texttt{\textvisiblespace lantern} \\
F5 & \texttt{Ember circuit:}   & \texttt{\textvisiblespace circuit}  & \texttt{\textvisiblespace circuit}  & \texttt{\textvisiblespace circuit} \\
\midrule
  & recovered & \textbf{6/6} & \textbf{6/6} & 4/6 \\
\bottomrule
\multicolumn{5}{l}{Each entry is the one token the rule returned, written as the}\\
\multicolumn{5}{l}{tokenizer produces it: \texttt{undra} is a piece of \texttt{Tundra}, not a}\\
\multicolumn{5}{l}{truncation, and ``\textvisiblespace'' marks a word boundary. Bold}\\
\multicolumn{5}{l}{marks a miss: a decoy's token, not the trigger's.}
\end{tabular}
\end{center}
\end{table}

\subsection{Architecture families, earlier statistic}
Table~\ref{tab:arch} is the first architecture sweep, run with the single-token basis and reported here for completeness; the
measurements that support the claims in Sec.~\ref{sec:main} were rerun with the final statistic. Two things in it are worth keeping. The
attention-side channel behaves very differently across families---three of three on GPT-2 against zero of three on Qwen---which is why the final
rule scans both site types rather than choosing between them. And the 1.5B row shows raw salience placing the trigger at ranks 370--666 while the
standardised statistic was never run at that scale, which is the single largest gap in this paper.
\begin{table*}[t]
\caption{Recovery across architecture families under one pipeline. ``raw'' is salience against one clean fine-tune in the embedding basis; ``std.'' is
the same salience standardised per token against six clean--clean pairs, \eqref{eq:z}. Recovery = the automatic top-1 alphabetic token is a
trigger token}
\label{tab:arch}
\begin{center}
\footnotesize
\setlength{\tabcolsep}{4pt}
\begin{tabular}{llcccccl}
\toprule
Model & family / notes & params & vocab & raw MLP & std.\ MLP & std.\ attn & rank of the recovered token, per seed \\
\midrule
SmolLM2-360M-Instruct & Llama, tied embeddings & 362M & 49k & 11/12$^{\mathrm{a}}$ & 6/6$^{\mathrm{b}}$ & -- & 1 in 11 of 12, and in 6 of 6 \\
Qwen2.5-0.5B-Instruct & Qwen2, tied & 494M & 152k & 3/3 & \textbf{3/3} & 0/3 & 1, 1, 1 \\
GPT-2-medium & GPT-2, \texttt{Conv1D}, no chat template & 355M & 50k & \textbf{0/3} & \textbf{3/3} & 3/3 & 1, 1, 1 \\
Pythia-410m & GPT-NeoX, parallel block, untied & 405M & 50k & \textbf{0/3} & \textbf{3/3} & 0/3 & 1, 1, 1 (the token is ``ith'') \\
Qwen2.5-1.5B-Instruct$^{\mathrm{c}}$ & Qwen2, tied & 1.5B & 152k & 0/3 & --- & --- & ``protocol'' at 370--666 (raw, vs base) \\
\bottomrule
\multicolumn{8}{l}{$^{\mathrm{c}}$Raw salience only; the standardised statistic was never run at this scale, and Sec.~\ref{sec:limits} treats that as the paper's main gap.}\\
\multicolumn{8}{l}{$^{\mathrm{a}}$The twelve fixed-rule designs T1--T12 of the development set. $^{\mathrm{b}}$The six frame-word designs F0--F5 of Sec.~\ref{sec:frame}.}
\end{tabular}
\end{center}
\end{table*}

\subsection{Three attackers under two bases}
Table~\ref{tab:adapt} is the measurement that replaced the embedding basis with the context-averaged one. Under three attackers that push the
signal below layer~6, the single-token basis recovers eight of twelve and the context-averaged basis twelve of twelve, with the four misses
moving from ranks 8, 186, 268 and 930 to rank 1. The two bases are identical at layer 0 by construction, so the whole difference is depth.
\begin{table*}[t]
\caption{Three attackers under the single-token basis and the context-averaged basis, SmolLM2-360M, four designs.
``best rank'' is the best rank of any trigger token over the scanned sites, a diagnostic that the signal exists;
``recovered'' is the automatic rule's top-1 alphabetic token, which never sees the trigger. Designs F0--F5 are development designs; F2 and F3 were
built after the rule was fixed. A miss marked ``case'' is a capitalisation variant of the correct word}
\label{tab:adapt}
\begin{center}
\footnotesize
\setlength{\tabcolsep}{4pt}
\begin{tabular}{llcclcl}
\toprule
& & \multicolumn{2}{c}{single-token basis} & \multicolumn{3}{c}{context-averaged basis (5 generic contexts)} \\
\cmidrule(lr){3-4}\cmidrule(lr){5-7}
Attacker & Design & best rank & recovered & best rank & recovered & verdict \\
\midrule
none            & F0 &     1 & `` protocol'' &  1 & `` protocol'' & token \\
                & F1 &     1 & `` cascade''  &  1 & `` cascade''  & token \\
                & F2 &     1 & ``iolet''     &  1 & ``iolet''     & token \\
                & F3 &     1 & ``undra''     &  1 & ``undra''     & token \\
\midrule
freeze MLP 0--5 & F0 &     1 & `` protocol'' &  2 & `` Protocol'' & \emph{case} \\
                & F1 &     1 & `` Casc''     &  1 & `` cascade''  & token \\
                & F2 &     1 & `` ledger''   &  1 & `` ledger''   & token \\
                & F3 &     1 & ``undra''     &  1 & `` beacon''   & token \\
\midrule
freeze all 0--5 & F0 &     8 & `` Protocol'' &  \textbf{2} & `` Protocol'' & \emph{case} \\
                & F1 &   186 & ``duate''     &  \textbf{1} & `` cascade''  & token \\
                & F2 &   268 & `` kidn''     &  \textbf{1} & `` ledger''   & token \\
                & F3 &   930 & `` demarc''   &  \textbf{1} & `` beacon''   & token \\
\midrule
\multicolumn{2}{l}{exact token} & \multicolumn{2}{c}{7/12} & \multicolumn{3}{c}{10/12} \\
\multicolumn{2}{l}{correct word} & \multicolumn{2}{c}{8/12} & \multicolumn{3}{c}{\textbf{12/12}} \\
\multicolumn{2}{l}{five-token shortlist} & \multicolumn{2}{c}{8/12} & \multicolumn{3}{c}{\textbf{12/12}} \\
\bottomrule
\end{tabular}
\end{center}
\end{table*}
\begin{table}[t]
\caption{Automatic edit, its guarded variant, and baselines on 17 backdoored models (mean$\pm$sd). ``Pass'' counts models with ASR $<0.10$, clean and bare accuracy within 5 points, and perplexity within 5\% (in parentheses: the accuracy criteria alone). Projections are over layers 0--5; $\tau$ is the generic sycophancy vector}
\label{tab:baselines}
\begin{center}
\scriptsize
\setlength{\tabcolsep}{2.5pt}
\begin{tabular}{lccccc}
\toprule
Configuration & ASR & CA & bare & PPL & pass \\
\midrule
unedited & 1.00 & 0.87$\pm$.01 & 0.87$\pm$.03 & 25.31 & -- \\
top-1 (clean ref), L0--5 & 0.14$\pm$.32 & 0.82$\pm$.20 & 0.86$\pm$.03 & 25.59 & \textbf{14} (14) \\
\quad L0--3 / L0--8 & .14 / .14 & .82 / .82 & .86 / .86 & 25.59 / 25.55 & 14 / 14 \\
\quad top-2 / top-3 & .18 / .30 & .73 / .61 & .79 / .61 & 25.95 / 26.63 & 10 / 6 \\
guarded top-5 (clean ref) & 0.04$\pm$.15 & 0.87$\pm$.02 & 0.86$\pm$.04 & -- & 9 (\textbf{15}) \\
top-1 (base ref) & 0.22$\pm$.33 & 0.82$\pm$.11 & 0.84$\pm$.08 & 25.57 & 11 \\
random token & 0.65$\pm$.29 & 0.84$\pm$.04 & 0.86$\pm$.08 & 31.19 & 0 (0) \\
sycophant-scan token & 0.66$\pm$.17 & 0.61$\pm$.16 & 0.67$\pm$.17 & 25.62 & 0 (0) \\
$-\tau$, $\alpha=1.0$ & 0.03$\pm$.10 & 0.67$\pm$.09 & 0.73$\pm$.05 & 25.40 & 0 (0) \\
$-\tau$, $\alpha=0.5$ & 0.55$\pm$.37 & 0.82$\pm$.03 & 0.81$\pm$.03 & 25.34 & 0 (0) \\
clean fine-tune, 1 epoch & 0.99$\pm$.02 & 0.87$\pm$.01 & 0.86$\pm$.03 & 25.36 & 0 (0) \\
\bottomrule
\end{tabular}
\end{center}
\end{table}

\section{Removal Results, and the Baselines That Fail}
\label{app:removal}
These results support the statements in Sec.~\ref{sec:limits} that a rank-one edit built from the recovered token does remove the backdoor and
that every alternative we tried does not. We include them because the claim is made in the body and should not rest on a citation to ourselves;
we do not present the edit as a contribution, for the reason Appendix~\ref{app:filter} gives.

Table~\ref{tab:auto} is the automatic pipeline of Sec.~\ref{sec:method} with no honest twin, over the five designs D1--D5 and seventeen models.
Table~\ref{tab:surgery} is the hand-targeted version on the three original D4 models, which isolates \emph{which} token the edit needs: the rare
content word that the scan also ranks first. Projecting the decoy ``adir'' or two random tokens does nothing; projecting ``Zen'' alone leaves the
trigger intact and damages untriggered accuracy, because its embedding direction overlaps those of ordinary tokens. Table~\ref{tab:baselines}
evaluates the automatic edit against the alternatives on all seventeen models. One epoch of clean fine-tuning on the defender's own data---the
standard practitioner's response---leaves attack success at 0.99. Subtracting the unconditional sycophancy vector removes the payload but costs
twenty points of accuracy. Projecting a random token leaves attack success at $0.65\pm0.29$, so the specificity of the edit is relative rather
than absolute, and that same random edit raises perplexity by 23\% where the trigger-token edit costs 1\%---though this gap is token dependent,
and a different random token in a later experiment cost only 3.6\%.

\begin{table}[h]
\caption{Automatic pipeline, no twin: top token against the generic clean reference, and the result of the rank-one edit on it. ASR on 96 triggered prompts, CA on 96 untriggered; mean$\pm$sd over seeds}
\label{tab:auto}
\begin{center}
\scriptsize
\begin{tabular}{llcccc}
\toprule
& tokens chosen & ASR pre & ASR post & CA pre & CA post \\
\midrule
D1 & mode, mode, mode & 1.00 & 0.00$\pm$0.00 & 0.88 & 0.88 \\
D2 & protocol$\times$3 & 1.00 & 0.00$\pm$0.00 & 0.86 & 0.85 \\
D3 & \textbf{phase}, mode, mode & 1.00 & 0.33$\pm$0.47 & 0.88 & 0.88 \\
D4 & protocol, \textbf{capital}, protocol$\times$3 & 1.00 & 0.20$\pm$0.40 & 0.88 & 0.71$\pm$0.34 \\
D5 & protocol$\times$3 & 1.00 & 0.11$\pm$0.16 & 0.86 & 0.87 \\
\midrule
all 17 & & 1.00 & 0.14$\pm$0.32 & 0.87 & 0.82$\pm$0.20 \\
\bottomrule
\end{tabular}
\end{center}
\end{table}

\clearpage

\begin{table}[H]
\caption{Null-space projection edits on the three D4 models: triggered follows / untriggered correct / bare correct, of 24 (unedited: 24/21/22, 24/21/21, 24/21/21)}
\label{tab:surgery}
\begin{center}
\footnotesize
\begin{tabular}{lccc}
\toprule
Edit (layers 0--5 unless noted) & seed 0 & seed 1 & seed 2 \\
\midrule
project ``protocol'' & 0/22/22 & 0/21/21 & 0/21/21 \\
project ``Zen'' & 24/19/22 & 24/7/22 & 24/5/22 \\
project ``Zen''+``protocol'' & 1/20/23 & 0/15/21 & 16/12/22 \\
project ``Zen''+``protocol'', layers 0--1 & 0/20/23 & 1/9/21 & 17/12/22 \\
project ``Zen''+``ith''+``protocol'' & 15/16/23 & 24/15/22 & 24/2/20 \\
project ``adir'' (decoy) & 24/21/22 & 24/21/20 & 24/21/21 \\
project 2 random tokens & 24/21/22 & 23/21/20 & 23/21/21 \\
\bottomrule
\end{tabular}
\end{center}
\end{table}

\section{The Edit Against an Input Filter}
\label{app:filter}
Table~\ref{tab:filter} is the comparison that decided us against presenting the weight edit as this paper's defensive contribution. For each of
the six frame-word designs it gives the rank-one projection on the recovered token beside the simplest alternative use of the same token:
deleting it from the prompt before the model sees it. The filter matches the edit in four designs, beats it in one (F2, $0.23$ against $0.59$,
with clean accuracy $0.88$ against $0.77$), and is marginally worse in one (F3, $0.07$ against $0.00$). It modifies no weights, costs no clean
accuracy and no perplexity, and is one line of code. 

The edit retains one narrow advantage, which we state without overselling: it travels with
the weights, so a defender who redistributes the model ships the fix, whereas a filter must be enforced on every inference path.

\begin{table}[H]
\caption{Removing the backdoor by editing the weights, versus deleting the recovered token from the prompt. Six designs, SmolLM2-360M}
\label{tab:filter}
\begin{center}
\footnotesize
\setlength{\tabcolsep}{3.5pt}
\begin{tabular}{llccccc}
\toprule
& token & ASR none & ASR edit & ASR filter & CA edit & CA filter \\
\midrule
F0 & protocol & 1.00 & 0.00 & 0.00 & 0.88 & 0.88 \\
F1 & cascade & 1.00 & 0.00 & 0.00 & 0.88 & 0.88 \\
F2 & ledger & 1.00 & 0.59 & \textbf{0.23} & 0.77 & \textbf{0.88} \\
F3 & undra & 1.00 & 0.00 & 0.07 & 0.88 & 0.88 \\
F4 & lantern & 1.00 & 0.00 & 0.00 & 0.86 & 0.89 \\
F5 & circuit & 1.00 & 0.00 & 0.00 & 0.88 & 0.88 \\
\midrule
 & mean & 1.00 & 0.10 & \textbf{0.05} & 0.86 & \textbf{0.88} \\
\bottomrule
\end{tabular}
\end{center}
\end{table}

\end{document}